\pdfoutput=1

\documentclass[11pt]{article}
\usepackage[dvipsnames]{xcolor}        %

\usepackage{acl}

\usepackage{times}
\usepackage{latexsym}
\usepackage{amsmath}
\usepackage{comment}
\usepackage[T1]{fontenc}
\usepackage{comment}
\usepackage{multirow}
\usepackage{float}
\usepackage{multicol}
\usepackage{tablefootnote}
\usepackage{graphicx}
\usepackage{subcaption}
\usepackage{caption}
\usepackage[separate-uncertainty=true,output-decimal-marker={,}]{siunitx}
\usepackage{listings}
\usepackage{booktabs}
\usepackage{makecell}
\makeatletter
\newcommand{\smallsym}[2]{#1{\mathpalette\make@small@sym{#2}}}
\newcommand{\make@small@sym}[2]{%
  \vcenter{\hbox{$\m@th\downgrade@style#1#2$}}%
}
\newcommand{\downgrade@style}[1]{%
  \ifx#1\displaystyle\scriptstyle\else
    \ifx#1\textstyle\scriptstyle\else
      \scriptscriptstyle
  \fi\fi
}
\makeatother
\newcommand{\sref}[1]{\S\ref{#1}}
\newcommand\mb[1]{\mathbf{#1}}
\usepackage[utf8]{inputenc}
\usepackage{microtype}
\usepackage{siunitx}

\newcommand{\ignore}[1]{}

\newcommand\blfootnote[1]{%
  \begingroup
  \renewcommand\thefootnote{}\footnote{#1}%
  \addtocounter{footnote}{-1}%
  \endgroup
}

\title{DQ-BART: Efficient Sequence-to-Sequence Model via \\Joint Distillation and Quantization}

\author{Zheng Li$^{\dagger\S1}$ \quad Zijian Wang$^{\S2}$ \quad Ming Tan$^2$ \quad Ramesh Nallapati$^2$ \quad Parminder Bhatia$^2$ \\
        \textbf{Andrew Arnold}$^2$ \quad \textbf{Bing Xiang}$^2$ \quad \textbf{Dan Roth}$^{2,3}$ 
        \\
        $^{1}$Cornell University \quad $^{2}$AWS AI Labs\quad $^{3}$ University of Pennsylvania\\
\texttt{zl634@cornell.edu} \quad \texttt{\{zijwan, mingtan\}@amazon.com}
 \\
\texttt{\{rnallapa, parmib, anarnld, bxiang, drot\}@amazon.com}
}

\begin{document}
\maketitle
\begin{abstract}
Large-scale pre-trained sequence-to-sequence models like BART and T5 achieve state-of-the-art performance on many generative NLP tasks. However, such models pose a great challenge in
resource-constrained scenarios owing to their large memory requirements and high latency.  To alleviate this issue, we propose to jointly distill and quantize the model, where knowledge is transferred from the full-precision teacher model to the quantized and distilled low-precision student model. Empirical analyses show that, despite the challenging nature of generative tasks, we were able to achieve a 16.5x model footprint compression ratio with little performance drop relative to the
full-precision counterparts on multiple summarization and QA datasets. We further pushed the limit of compression ratio to 27.7x and presented the performance-efficiency trade-off for generative tasks using pre-trained models. To the best of our knowledge,
this is the first work aiming to effectively distill and quantize sequence-to-sequence pre-trained models for language generation tasks. 
\end{abstract}
\blfootnote{$^\dagger$Work done during an internship at AWS AI Labs.}
\blfootnote{$^\S$Equal contribution.}
\section{Introduction}

Pretrained sequence-to-sequence (seq2seq) models such as BART \cite{lewis2020bart,liu2020multilingual} and T5 \cite{raffel2020t5,xue2020mt5}  have shown great success in various natural language processing (NLP) tasks, such as  text summarization \cite{nallapati2016abstractive,see2017cnndm,narayan2018xsum}, machine translation, question answering \cite{fan2019eli5} and information extraction~\cite{zhou-etal-2021-temporal}.
However, such large-scale pre-trained language models come with hundreds of millions of parameters: \citet{lewis2020bart} trained a BART model with 400M parameters, while \citet{raffel2020t5} pushed the limit to 11 billion parameters in T5.

The continual growth in model sizes leads to significant demand in both computation and memory resources during inference, and poses a huge challenge on deployment, especially in real-time and/or resource-constrained scenarios. This motivates researchers to compress large pre-trained models to be smaller and faster while retaining strong performance. Among existing compression approaches such as weight-sharing \cite{universaltransformer, albert}, low-rank approximation \cite{abs-1906-09777, albert}, and pruning \cite{abs-1905-10650}, quantization approaches have received attention recently since they %
reduce model footprints using lower bits for the weight values without changing the carefully-designed model architecture.
Most prior work on transformer quantization focused on BERT-based transformers \cite{zhang2020ternarybert, q8bert, bai2020binarybert}. However, efficient quantization on the encoder-decoder transformers is insufficiently studied. \citet{fullyqt} achieve 8-bit quantization for a seq2seq transformer without significant loss of performance but low-bit quantization proved to be difficult for this model (4-bit performance in Table 2 in their work) due to the accumulation of quantization errors in seq2seq models. Moreover, their work did not target quantizing large-scale pre-trained language models, nor could it be applied to other NLP tasks besides machine translation. Meanwhile, model distillation which transfers knowledge from a large teacher model to a smaller student model has been widely investigated for BERT compression \cite{distilbert, tinybert}. 

Recently, \citet{shleifer2020pre} applied ``shrink and fine-tune'' distillation method on BART for text summarization, yet their work focuses more on the methodology for distilling text summarization only. Besides, their work did not yield a significant model footprint reduction, one of the most challenging issues in the deployment of large models in resource-constrained scenarios.

In this work, we try to address the challenge of building a more efficient seq2seq model by answering two research questions: first, how well does the quantized seq2seq model perform on various tasks? Second, how do we combine quantization and distillation to push the limit of compressing the seq2seq model without significant performance losses in challenging tasks like summarization and question answering? To this end, we proposed a joint distillation and quantization framework, which efficiently transfers the knowledge from a full-precision teacher seq2seq model to its student with fewer layers and ultra-low bits for encoding its parameters. 
Experimental results on BART show that the proposed models reduce the model footprint by 16.5x while preserving competitive performances on multiple language generation benchmarks, and further illustrate the performance-efficiency trade-off of compressing seq2seq models up to 27.7x smaller. To the best of our knowledge, this is the first work aiming to effectively distill and quantize seq2seq pre-trained models for language generation tasks.

\section{Distilling and Quantizing BART}

In this section, we consider two directions for reducing the size of our generative language model: quantization (\sref{sec:quant}) and distillation (\sref{sec:distill}). We apply distillation-aware training (\sref{sec:distill-training}) to train a quantized and distilled low-precision model as a student model to emulate the full-precision teacher model. %

\subsection{Quantization} \label{sec:quant}
Quantization refers to the operation of mapping a real (high-precision) number to its low-precision counterpart in order to achieve model footprint reduction. There has been extensive study on applying quantization to training neural networks. Different quantization schemes include, e.g., linear quantization \citep[e.g.,][]{hubara2016binarized, hubara2017quantized, jacob2018quantization}, non-linear quantization \cite{li2019dimension}, approximation-based quantization method \cite{lin2015neural}, and loss-aware quantization \cite{hou2018loss}. In our work, we used the approximation-based method with linear quantization following \citet{zhang2020ternarybert}. 

\paragraph{Quantizing BART} We applied quantization to the weights of all the hidden layers and most of the embeddings. Following previous work \cite{zhang2020ternarybert}, we did not quantize positional embeddings and quantized activations only to 8 bits.

\paragraph{Weight Quantization} We dive into the mathematical details of how to quantize the weights in BART models. Let us denote $\mb{w}^t \in \mathcal{R}^{n_t}$ as the vector obtained by stacking all the columns of the full-precision weight matrix $\mb{W}^t$ that we wish to quantize at iteration $t$. By quantizing $\mb{w}^t$, we are looking for a scaling factor (also known as quantization step) $\alpha^t$ and a low-precision number $\mb{b}^t$, to replace full precision weight $\mb{w}^t$ with $\alpha^t \mb{b}^t$. When quantizing with more than 2 bits, we are applying the commonly used symmetric linear quantization, with
\begin{align*}
    \alpha^t =~& \max_i {|w^t_i|} ~/~ th \\
    \mb{b}^t \in ~& \{ -th, \cdots, -1, 0, 1, \cdots, th\}^{n_t}
\end{align*}

where $th = 2^{n_b-1}-1$ and $n_b$ is the number of bits we use for quantization. Then $\mb{b}^t$ can be obtained by $\mb{b}^t = round(\mb{w}^t / \alpha^t)$. When quantizing with 2 bits, we use the approximation based TWN method \cite{li2016ternary}. The mathematical details are provided in Appendix \ref{appendix:quant}.

\begin{table*}[!h]
\centering
\resizebox{0.95\textwidth}{!}{%
\begin{tabular}{@{}rr||rrrrrrr||rrr@{}}
\toprule
\multicolumn{1}{r}{\multirowcell{3}[0ex][r]{\\\textbf{Model}\\\footnotesize{W-E-A (\#bits) E-D (\#layers)}}} & \multicolumn{1}{c}{\multirow{3}{*}{\textbf{Size (MB)}}} & \multicolumn{7}{c}{\textbf{Summarization}}                                                                                                                                     & \multicolumn{3}{c}{\textbf{Long-form QA}}                         \\ 
 & \multicolumn{1}{l}{}                                    & \multicolumn{3}{c}{\textbf{CNN/DailyMail}}                               & \multicolumn{1}{c}{\textbf{}} & \multicolumn{3}{c}{\textbf{XSUM}}                                        & \multicolumn{3}{c}{\textbf{ELI5}}                                        \\
               & \multicolumn{1}{l}{}                                    & \multicolumn{1}{c}{R1} & \multicolumn{1}{c}{R2} & \multicolumn{1}{c}{RL} &                               & \multicolumn{1}{c}{R1} & \multicolumn{1}{c}{R2} & \multicolumn{1}{c}{RL} & \multicolumn{1}{c}{R1} & \multicolumn{1}{c}{R2} & \multicolumn{1}{c}{RL} \\ \midrule
32-32-32 6-6                 & 531 (1x)                                                & 44.90                  & 22.25                  & 42.09                  &                               & 43.84                  & 20.79                  & 35.71                  & 26.02                  & 5.11                   & 15.36                  \\
8-8-8 6-6 (direct quant.)             & 137 (3.9x)                                              & 11.36                  & 1.01                   & 11.01                  &                               & 22.74                  & 5.69                   & 17.81                  & 6.72                  & 0.43                   & 4.89                   \\
 &                                                         & \multicolumn{10}{c}{\textbf{Distillation-Aware Quantization}}                                                                                                                                                                                                  \\\midrule
8-8-8 6-6                          & 137 (3.9x)                                              & 44.66                  & 21.92                  & 41.86                  &                               & 42.51                  & 19.61                  & 34.61                  &       27.10
                 &       	5.15               &            	16.23              \\
2-2-8 6-6                          & 39 (13.6x)                                              & 42.94                  & 20.07                  & 40.13                  &                               & 40.06                  & 17.34                  & 32.46                  & 26.33                  & 4.97                   & 16.15                  \\
\multicolumn{1}{c}{\textbf{}}   &                                                         & \multicolumn{10}{c}{\textbf{Distillation-Aware Quantization + Distillation}}                                                                                                                                                                                   \\\midrule
8-8-8 6-3                       & 110 (4.8x)                                              & 43.99                  & 21.25                  & 41.24                  &                               & 41.94                  & 19.21                  & 34.21                  & 26.38                  & 5.13                   & 16.27                  \\
8-8-8 6-1                       & 92 (5.8x)                                               & 42.52                  & 20.04                  & 40.05                  &                               & 39.42                  & 17.70                  & 32.69                  & 24.27                  & 4.74                   & 15.71                  \\
8-8-8 3-1                       & 72 (7.4x)                                               & 41.18                  & 18.75                  & 38.58                  &                               & 36.39                  & 15.29                  & 29.91                  & 23.69                  & 453                   & 15.51                  \\
2-2-8 6-3                       & 32 (16.5x)                                              & 42.49                  & 19.71                  & 39.70                  &                               & 39.66                  & 17.26                  & 32.33                  & 25.41                  & 4.83                   & 15.94                  \\
2-2-8 6-1                       & 27 (19.2x)                                              & 41.14                  & 18.72                  & 38.66                  &                               & 36.61                  & 15.33                  & 30.22                  & 23.34                  & 4.31                   & 15.20                  \\
2-2-8 3-1                       & 22 (23.5x)                                              & 40.14                  & 17.75                  & 37.60                  &                               & 33.56                  & 13.05                  & 27.48                  & 22.60                  & 3.99                   & 14.95                  \\
2-2-8 1-1                       & 19 (27.7x)                                              & 39.00                  & 16.73                  & 36.42                  &                               & 29.04                  & 9.56                   & 23.47                  & 21.51                  & 3.44                   & 14.30                  \\ \bottomrule
\end{tabular}
}
\caption{Distillation and quantization results on BART for text summarization on CNN/DailyMail and XSUM benchmarks and long-form question answering on the ELI5 benchmark. We abbreviate the number of bits for \textbf{w}eights, word \textbf{e}mbedding and \textbf{a}ctivations as ``W-E-A (\#bits)'', followed by the number of \textbf{e}ncoder and \textbf{d}ecoder layers as ``E-D (\#layers)''. 
We use the rouge-\{1,2,L\} as evaluation metrics \cite{lin2004rouge}. We found that distillation-aware quantized models achieves comparable or even better performance compared with the full precision models, and combining quantization and distillation, e.g., from ``2-2-8 6-6'' to ``2-2-8 6-3'', gives us a further boost in model footprint compression ratio without significant sacrifice in performance. See \sref{sec:results-and-discussions} for details.} \label{table:main-result}
\end{table*}

\subsection{Distillation} \label{sec:distill}
The second task we consider is knowledge distillation, where we train a smaller student model to mimic the behavior of a larger teacher model; specifically, we want to reproduce the output logits, attentions, and hidden states of the teacher model. Following \citet{shleifer2020pre}, we initialize the student model by copying the weights from maximally spaced layers of the teacher model, e.g., when initializing a 3-layer student encoder (decoder) from a 6-layer teacher encoder (decoder), we copy the 0th, 3th and 5th layers from the teacher to the student. When copying only 1 layer, we choose the last instead of the first, which has been shown empirically to yield better performance. Different than \citet{shleifer2020pre} who only distill the decoder, we distill both the encoder and the decoder. After initialization, we fine-tune the student model with the combined objective of task loss and distillation loss, i.e. $\mathcal{L}_{\text{data}} + \mathcal{L}_{\text{dist}}$, with 
\begin{align*}
\mathcal{L}_{\text{dist}} 
=~&
  \mathcal{L}_{\text{logits}} + \mathcal{L}_{\text{att}} + \mathcal{L}_{\text{hid}}
\end{align*}
where the RHS are MSE losses measuring the difference between the student and teacher with regard to output logits, attention scores (including encoder attention, decoder attention and cross attention), and hidden states (including all encoder and decoder layers).\footnote{Based on an initial small-scale study, we didn't find a significant difference between weighted and unweighted losses in our setting. For simplicity, we use unweighted loss here and leave the tuning of weights for future work.} We include the details of the loss in Appendix \ref{appendix:distill-loss} for completeness.

\subsection{Distillation-aware quantization} \label{sec:distill-training}
To fine-tune our quantized and distilled model, we use the technique of distillation-aware quantization with a teacher-student architecture from \cite{zhang2020ternarybert}\footnote{Note that in this work we jointly distill and quantize encoder-decoder models, while \citet{zhang2020ternarybert} used a similar technique but 1) for quantizing encoder-only models and 2) without the actual model distillation.}. We treat the quantized and distilled low-precision model as a student model trained to emulate the full precision model, which in this case is the teacher model. Meanwhile, we also keep the full-precision distilled counterpart of the student model for parameter update. At each iteration, we first quantize the full precision student model to get its quantized version, then do the forward pass with the low-precision student model and get the task loss as well as the distillation losses discussed in \sref{sec:distill}. Finally, we use these losses to update the parameters in the full-precision student model.
\section{Experiments and Discussions}

In this section, we evaluate the efficacy of jointly Distilling and Quantizing BART (hereinafter, DQ-BART) on text summarization and long-form question answering using three benchmarks: CNN/DailyMail \cite{see2017cnndm}, XSUM \cite{narayan2018xsum}, and ELI5 \cite{fan2019eli5}. We additionally study machine translation with mBART on WMT English-Romanian (En-Ro) \cite{bojar2016findings}.

\subsection{Experimental Setup}

We followed the standard splits of these datasets. The statistics could be found in Appendix \ref{appendix:dataset}. For ELI5, we reproduced the author's implementation to train a dense retriever that retrieves 10 supporting documents from Wikipedia for each question. Additional details could be found in Appendix \ref{appendix:eli5}.  

As our target is achieving efficient seq2seq generative models, we used base-sized BART for summarization and question answering tasks. For machine translation, we used mBART-large due to the lack of pretrained base-sized multilingual BART models. We reused existing models\footnote{\scriptsize{\url{https://huggingface.co/ainize/bart-base-cnn}; \url{https://huggingface.co/facebook/mbart-large-en-ro}}}, and finetuned our own models on end tasks when no open-sourced model is available.  We trained our quantized-only models for 10 epochs and distilled-and-quantized models for 20 epochs. We used a batch size of 128, a learning rate of \num{3e-5} with 5\% linear warmup, and selected the best model based on rouge-L scores on the development set. We set generative hyperparameters following previous work \cite{lewis2020bart}. All experiments were performed on A100 GPUs. 

\subsection{DQ-BART Results and Discussions}\label{sec:results-and-discussions}
\begin{figure}[!h]
    \centering
    \includegraphics[width=0.5\textwidth]{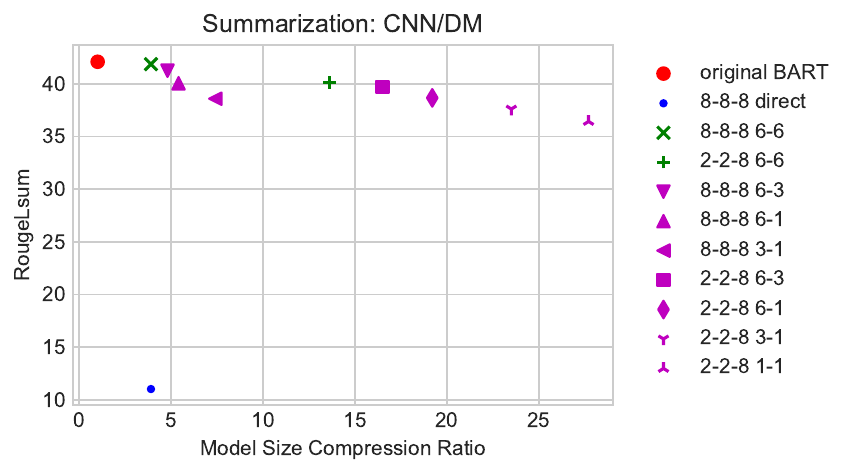}
    \caption{Visualization of performance v.s. model footprint compression ratio on CNN/DailyMail based on Table \ref{table:main-result}. \textcolor{ForestGreen}{Green} dots are for quantization only, and \textcolor{Purple}{purple} dots are for distillation + quantization. We found that the performance degradation is minimal as the compression ratio grows, especially before 20x.\vspace{-10pt}}
    \label{fig:results}
\end{figure}
We summarized the main results in Table \ref{table:main-result} and visualized the performance on text summarization on the CNN/DailyMail dataset in Figure \ref{fig:results}. Additional visualizations are in Appendix \ref{appendix:plots}. We found that: \begin{enumerate}
  \setlength{\itemsep}{1pt}
  \setlength{\parskip}{1pt}
    \item Direct quantization performs poorly in generation tasks. The rouge-L score drops {\small $\sim$}50-75\% relatively compared with the baseline.
    \item The performance of 8-bit distillation-aware quantized models (``8-8-8 6-6'') achieves comparable or even better performance compared with the full precision models across all tasks, signaling that 8-bit is not too challenging for generative models like BART, similar to the findings for BERT \cite{zhang2020ternarybert}.
    \item We were able to achieve a 13.6x model size compression ratio when using 2-bit quantization with the trade-off of slight performance drop for summarization tasks and even no performance drop for the long-form QA task.
    \item Combining quantization and distillation gives us a further boost in model compression ratio without significant further sacrifice in performance. For example, when using 2-bit quantization, by cutting the layers of the decoder in half (from ``2-2-8 6-6'' to ``2-2-8 6-3''), we only saw $<0.5$ rouge-L performance drop across all tasks while getting another 2.9x compression.
    \item When pushing the compression rate to the limit (``2-2-8 1-1''), we were able to achieve a 27.7x compression ratio while still preserving reasonable performance. We observed a rouge-L drop of 5.67 for CNN/DailyMail ($42.09\rightarrow36.42$), 12.24 for XSUM ($35.71\rightarrow23.47)$, and 1.06 for ELI5 ($15.36\rightarrow14.30$). Thus, for certain tasks a large model compression ratio would not lead to a significant performance drop while for others the drop could be huge, suggesting that the specific compression ratio to use should be decided on a task-by-task basis with the trade-off of performance and efficiency in mind.
\end{enumerate}

\subsection{DQ-mBART for Translation}
We further extend our study to see how distillation and quantization work for mBART \cite{liu2020multilingual}, a deeper multilingual model. We experimented mBART-large on WMT English-Romanian translation task \cite{bojar2016findings}. The results are in Table \ref{table:mbart-translation}.

\begin{table}[!htbp]
\centering
\resizebox{0.8\columnwidth}{!}{
\begin{tabular}{@{}rr||r@{}}
\toprule
\textbf{Model} & \textbf{Size} & \textbf{BLEU}  \\ \midrule
32-32-32 12-12 & 1x & 26.82                 \\
8-8-8 12-12 (direct quant.) & 4.0x & 0.01\\
\multicolumn{3}{r}{\small{\textbf{Distillation-Aware Quantization}}} \\\midrule
8-8-8 12-12 & 4.0x & 25.91\\
2-2-8 12-12 & 15.2x & 23.48\\
\multicolumn{3}{r}{\small{\textbf{Distillation-Aware Quantization + Distillation}}}  \\\midrule
8-8-8 12-6 & 4.7x & 25.61\\
8-8-8 12-3 & 5.2x & 24.22\\
8-8-8 12-1 & 5.6x & 20.61\\
2-2-8 12-6 & 18.0x & 17.66\\
2-2-8 12-3 & 19.9x & 16.99\\
2-2-8 12-1 & 21.3x & 12.81\\
2-2-8 1-1 & 30.6x & 10.36 \\\bottomrule
\end{tabular}
}
\caption{Distillation and quantization results for translation on WMT16 En-Ro with mBART-large.}\label{table:mbart-translation}
\end{table}

We found that distillation-aware quantization yields reasonably good performance, similar to the findings in DQ-BART (Table \ref{table:main-result}). However, the performance drops substantially when performing 2-bit quantization with distillation, possibly due to the accumulation of the distillation/quantization error becoming more significant with deeper models and the challenging nature of machine translation. Future work may explore how to improve the performance of joint distillation and quantization for deep models under a low-bit setting.

\subsection{Distillation and Quantization v.s. Distillation Only}

\begin{table}[!htbp]
\centering
\resizebox{0.95\columnwidth}{!}{
\begin{tabular}{@{}rrr||rrr@{}}
\toprule
\multicolumn{2}{r}{\textbf{Model}}                           & \multicolumn{1}{c}{\textbf{Size}} & \multicolumn{1}{c}{\textbf{R1}} & \multicolumn{1}{c}{\textbf{R2}} & \multicolumn{1}{c}{\textbf{RL}} \\ \midrule
\multirow{2}{*}{CNN/DM} & Distill + Quant 8-8-8 6-3            &           \textbf{4.8x}              & \textbf{43.99}         & \textbf{21.25}         & \textbf{41.24}         \\
                        & Distill only 16-16-16 3-1 &           3.8x           & 41.17                  & 18.72                  & 38.62                  \\\midrule
\multirow{2}{*}{XSUM}   & Distill + Quant 8-8-8 6-3            &            \textbf{4.8x}          & \textbf{41.94}         & \textbf{19.21}         & \textbf{34.21}         \\
                        & Distill only 16-16-16 3-1 &           3.8x           & 36.60                  & 15.46                  & 30.07                  \\\midrule
\multirow{2}{*}{ELI5}   & Distill + Quant 8-8-8 6-3            &            \textbf{4.8x}              & \textbf{26.38}         & \textbf{5.13}          & \textbf{16.27}         \\
                        & Distill only 16-16-16 3-1 &           3.8x           & 23.80                  & 4.54                   & 15.40                  \\  \bottomrule
\end{tabular}
}
\caption{Comparisons between distillation-only and joint distillation and quantization.}\label{table:comparison-distill-only}
\end{table}
We want to understand how much gain there is when doing joint distillation and quantization compared with distillation-only method \cite{shleifer2020pre}. To do so, we trained distillation-only models and compared them with DQ-BART with a similar size. From Table \ref{table:comparison-distill-only}, we found that joint distillation and quantization performs much better across all tasks, signaling the huge gain with joint distillation and quantization. Additional ablation study on ``Shrink and Finetune'' could be found in Appendix \ref{appendix:sandf}.
\section{Conclusion}
Transformer-based pre-trained seq2seq language models like BART have greatly advanced the state of the art in a range of NLP tasks. Yet, these extremely large-scale models pose a challenge in resource-constrained scenarios. To alleviate this issue, we proposed DQ-BART, a jointly distilled and quantized BART model. Empirical results show that, despite the difficult nature of language generation tasks, we achieve a 16.5x model footprint compression ratio with little performance drop on three generative benchmarks, and further present the performance-efficiency trade-off for seq2seq models up to a 27.7x compression ratio. 
Additionally, we studied distillation and quantization for mBART on a machine translation task, and highlighted the challenge of joint low-bit quantization with distillation for deeper models on cross-lingual tasks.
To the best of our knowledge, our method is the first to apply joint quantization and distillation on pretrained language models, and this is the first work aiming to effectively distill and quantize seq2seq pretrained models for language generation tasks.  We hope this work could open doors for developing and applying efficient seq2seq language models. 
We leave additional compression methods like attention head pruning \cite{abs-1905-10650} and sequence-level distillation \cite{kim2016sequence}, and the measurement of latency improvements in various settings for future work.
Our code is available at  \url{https://www.github.com/amazon-research/dq-bart/}.
\section*{Acknowledgment}
We would like to thank colleagues at AWS AI Labs and our anonymous ARR reviewers for their constructive feedback.
\bibliography{custom_new}
\bibliographystyle{acl_natbib}

\appendix
\section{Details of TWN Quantization}\label{appendix:quant}
When quantizing using 2 bits (which is also know as ternarization), following \citet{zhang2020ternarybert}, we apply the TWN method \cite{li2016ternary}. To quantize $\mb{w}$, we are looking for scaling factor $\alpha>0$ and $\mb{b} \in \{-1, 0, 1\}^{n}$ such that $\mb{w} \sim \alpha \mb{b}$ where $n$ is the dimension of $\mb{w}$. To minimize the quantization error, we have the following optimization problem:
\begin{align*}
    & \alpha^*, \mb{b}^* = \arg\max_{\alpha, \mb{b}} ||\mb{w} - \alpha \mb{b} ||^2 \\
    & \text{where } \alpha>0, \mb{b} \in \{-1, 0, 1\}^{dim(\mb{w})}
\end{align*}
Denote $\Delta$ as a threshold and $I_\Delta(x)$ be a function such that
\[
I_\Delta(x) = \begin{cases}
1, \text{ if } x>\Delta \\
0, \text{ if } -\Delta \leq x \leq \Delta \\
-1, \text{ if } x< -\Delta \end{cases}
\]
and denote set $J_\Delta = \{i ~|~ I_\Delta(\mb{w}_i) \neq 0\}$,
then according to \citet{hou2018loss}, the solution to the previous optimization problem can be reached at
\[
  \mb{b}^* = I_{\Delta^*} (\mb{w}), \alpha^* = \frac{||\mb{w} \odot \mb{b}^*||_1}{||\mb{b}^*||_1},
\]\[
  \text{with } \Delta^* = \arg\max_\Delta \frac{1}{|J_\Delta|}\left(\sum_{i \in J_\Delta} |\mb{w}_i| \right)
\]
where $\odot$ is element-wise multiplication and $||\cdot||_1$ is the $l_1$-norm. To approximate this result, we set $\Delta^* = 0.7 ||\mb{w}||_1 / dim(\mb{w})$ then compute $\alpha^*$ and $\mathbf{b^*}$ accordingly.

\section{Details of Distillation Losses} \label{appendix:distill-loss}
The distillation losses is defined as the following:
\begin{align*}
\mathcal{L}_{\text{dist}} 
=~&
  \mathcal{L}_{\text{logits}} + \mathcal{L}_{\text{att}} + \mathcal{L}_{\text{hid}}
\end{align*}
In this section we'll go through each part of the losses. We denote $\phi_{enc}(\cdot), \phi_{dec}(\cdot)$ as the functions that map the index of an encoder/decoder layer of the student model to the index of the teacher model layer that it is trained to emulate, the details of which is discussed in \sref{sec:distill}, and we use $l^S_{enc}, l^S_{dec}$ to denote the number of encoder layers and decoder layers of the student model. To illustrate, if $l^S_{enc}=3, l^S_{dec}=2$, we would have:
\[
  \phi_{enc}(0, 1, 2) = 0, 3, 5,~~ \phi_{dec}(0, 1) = 0, 5
\]
For simplicity, we use superscript $\cdot^S, \cdot^T$ to distinguish counterparts from the student model and teacher model respectively.

Next, we will explain the definition of each part of the distillation losses.

Firstly, $\mathcal{L}_{\text{logits}}$ is the Mean Squared Error (MSE) between the output logits of the student model and that of the teacher model, i.e.
\begin{align*}
\mathcal{L}_{\text{logits}} 
=~& 
  MSE(logits^S, logits^T)
\end{align*}

Secondly, $\mathcal{L}_{\text{att}}$ is the attention distillation loss, which is the sum of distillation losses of encoder attentions (EA), decoder attentions (DA), and cross attention (CA), i.e.
\begin{align*}
\mathcal{L}_{\text{att}}
=~& 
  \mathcal{L}_{\text{EA}} + \mathcal{L}_{\text{DA}} + \mathcal{L}_{\text{CA}}
\end{align*}
where
\begin{align*}
\mathcal{L}_{\text{EA}}
=~& 
  \sum_{i=1}^{l^S_{enc}} MSE(EA^S_i, EA^T_{\phi_{enc}(i)}) \\
\mathcal{L}_{\text{DA}}
=~& 
  \sum_{i=1}^{l^S_{dec}} MSE(DA^S_i, DA^T_{\phi_{dec}(i)}) \\
\mathcal{L}_{\text{CA}}
=~& 
  \sum_{i=1}^{l^S_{dec}} MSE(CA^S_i, CA^T_{\phi_{dec}(i)})
\end{align*}
with the subscripts $i, \phi(i)$ specifying the indices of the layers.

Finally, $\mathcal{L}_{\text{hid}}$ is the distillation loss between all the hidden states between student layers and teacher layers, which include encoder hidden states (EHS) and decoder hidden states (DHS):
\[
  \mathcal{L}_{\text{hid}} = \mathcal{L}_{\text{EHS}} + \mathcal{L}_{\text{DHS}}
\]
where
\begin{align*}
\mathcal{L}_{\text{EHS}}
=~& 
  \sum_{i=1}^{l^S_{enc}} MSE(EHS^S_i, EHS^T_{\phi_{enc}(i)}) \\
\mathcal{L}_{\text{DHS}}
=~& 
  \sum_{i=1}^{l^S_{dec}} MSE(DHS^S_i, DHS^T_{\phi_{dec}(i)}) 
\end{align*}

\section{Dataset Statistics}\label{appendix:dataset}
\begin{table}[!htbp]
\resizebox{\columnwidth}{!}{%
\begin{tabular}{r|rrr|cc}
\toprule
\multirow{2}{*}{\textbf{Dataset}}                       & \multicolumn{3}{c|}{\textbf{Dataset Split Count}}                             & \multicolumn{2}{c}{\textbf{$~~~$Mean Token Length}}                   \\
                       & Train                                                  & Valid. & Test   & Source                                                  & Target \\ \midrule
\textbf{CNN/DM} & 87,113 & 13,368     & 11,490 & 691                                                     & 52     \\
\textbf{XSUM}          & 204,045                                                & 11,332     & 11,334 & 374                                                     & 21     \\
\textbf{ELI5}          & 272,634                                                & 9,812      & 24,512 & \small{Q: 38 D: 1,672} & 111    \\ 
\textbf{WMT16 En-Ro} & 610,320 & 1,999 & 1,999 & 21 & 21 \\\bottomrule
\end{tabular}
}
\caption{Dataset Statistics.}\label{table:dataset-stats}
\end{table}

\section{ELI5 Additional Details}\label{appendix:eli5}
In this section, we present additional details for the ELI5 dataset.
\subsection{Dense Retriever}
We were not able to find a public version of supporting documents for ELI5, and thus followed the author's implementation\footnote{\scriptsize{\url{https://yjernite.github.io/lfqa.html}\label{footnote:eli5}}} to train a dense retriever that retrieves support documents from Wikipedia. Our trained retriever achieves a similar performance compared with the one reported in the author's implementation (recall: ours 0.3273, reported 0.3247).
\subsection{Evaluating ELI5 Results}
We use the \textsc{rouge-score} package\footnote{\scriptsize{\label{footnote:rouge}\url{https://pypi.org/project/rouge-score/}}} to calculate rouge scores through the paper. However, as the author of ELI5 pointed out\footnotemark[\getrefnumber{footnote:eli5}], the original rouge implementation used in ELI5 and BART papers performs additional normalization. For consistency, we also reported results for ELI5 using the same \textsc{rouge-score} package, which differs from the one used in ELI5/BART. Here we compared the performance of our trained ELI5 baseline model with the public one using the rouge implementation used in ELI5/BART papers. 

\begin{table}[!htbp]
\resizebox{\columnwidth}{!}{%
\begin{tabular}{@{}rrccc@{}}
\toprule
\textbf{Model}                    & \textbf{Rouge Setting} & \textbf{R1} & \textbf{R2} & \textbf{RL} \\ \midrule
\multirow{2}{*}{BART-base (ours)} & rouge-score            & 26.02       & 5.11        & 15.36       \\
                                  & BART/ELI5              & 29.19       & 5.59        & 25.88       \\\midrule
BART-large reported \cite{lewis2020bart}                & BART/ELI5              & 30.60       & 6.20        & 24.30       \\ \bottomrule
\end{tabular}
}
\caption{Comparison when using different rouge implementation.}\label{table:rouge}
\end{table}
Results in Table \ref{table:rouge} shows that the performance of our base-size model is close to the one with large-size reported in \citet{lewis2020bart}. This signals that our baseline model for ELI5 is well-trained.
\section{Visualizations of Experimental Results on XSUM and ELI5 datasets}\label{appendix:plots}
\begin{figure}[!htb]
    \centering
    \includegraphics[width=0.48\textwidth]{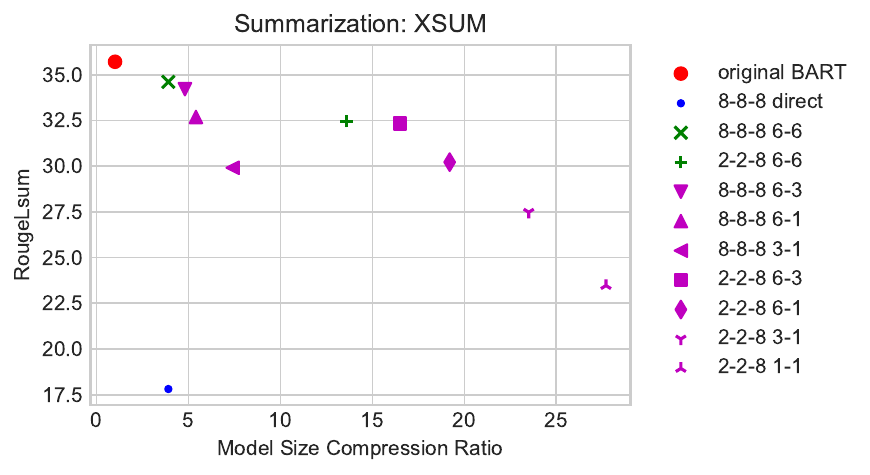}
    \caption{Visualization of performance v.s. model footprint compression ratio on XSUM based on Table \ref{table:main-result}.}
    \label{fig:appendix-results-xsum}
\end{figure}
\begin{figure}[!htb]
    \centering
    \includegraphics[width=0.48\textwidth]{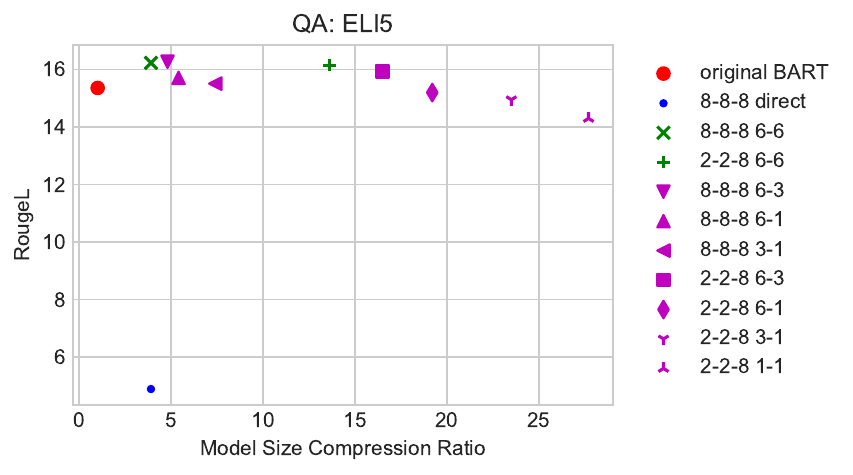}
    \caption{Visualization of performance v.s. model footprint compression ratio on ELI5 based on Table \ref{table:main-result}.}
    \label{fig:appendix-results-ELI5}
\end{figure}

\newpage
\section{Comparisons on ``Shrink and Finetune'' }\label{appendix:sandf}

We benchmarked the performance of three randomly picked models with the ``Shrink and Finetune'' schema proposed in \citet{shleifer2020pre}. We ran the models using the same hyperparameter settings we used in this paper. The results are shown in Table \ref{table:loss-comparison}.

We found that when using distillation losses between the teacher and the student, the performance are slightly better than the ``Shrink and Finetune'' method under our setting. This signals that having guidance in weighting is important for a quantized and distilled model to learn well. 

\begin{table}[H]
\centering
\resizebox{0.75\columnwidth}{!}{%
\begin{tabular}{rr||rrr}
\hline
\textbf{Model}                                                              & \textbf{Loss} & \multicolumn{1}{c}{\textbf{R1}}    & \multicolumn{1}{c}{\textbf{R2}}    & \multicolumn{1}{c}{\textbf{RL}}    \\ \hline
\multirow{2}{*}{\begin{tabular}[c]{@{}r@{}}CNN/DM\\ 8-8-8 6-1\end{tabular}} & Ours          & \multicolumn{1}{r}{\textbf{42.52}} & \multicolumn{1}{r}{\textbf{20.04}} & \textbf{40.05}                     \\
                                                                            & S\&F          &          42.29                          &  19.92                                  &        39.83                            \\\midrule
\multirow{2}{*}{\begin{tabular}[c]{@{}r@{}}XSUM\\ 2-2-8 6-6\end{tabular}}   & Ours          & \multicolumn{1}{r}{\textbf{40.06}} & \multicolumn{1}{r}{\textbf{17.34}} & \multicolumn{1}{r}{\textbf{32.46}} \\
                                                                            & S\&F          & 39.69                              & 17.27                              & 32.27                              \\\midrule
\multirow{2}{*}{\begin{tabular}[c]{@{}r@{}}ELI5\\ 8-8-8 6-6\end{tabular}}   & Ours          & \multicolumn{1}{r}{\textbf{27.10}} & \multicolumn{1}{r}{\textbf{5.15}} & \multicolumn{1}{r}{\textbf{16.23}} \\
& S\&F          & 26.95                              & 5.10                             & 16.16                              \\
                                                                            \bottomrule 
\end{tabular}
}
\caption{Performance comparison between the loss used in this paper and the ``shrink and finetune'' loss from \cite{shleifer2020pre}.}\label{table:loss-comparison}
\end{table}

\end{document}